\documentclass{article}
\pdfoutput=1
\PassOptionsToPackage{numbers, compress}{natbib}

\usepackage[preprint]{neurips_2024}

\usepackage[utf8]{inputenc} 
\usepackage[T1]{fontenc}    
\usepackage{hyperref}       
\usepackage{url}            
\usepackage{booktabs}       
\usepackage{amsfonts}       
\usepackage{nicefrac}       
\usepackage{microtype}      
\usepackage{xcolor}         

\usepackage{amsthm}

\newtheorem{proposition}{Proposition}
\newtheorem{fact}{Fact}

\usepackage{cite}
\usepackage{enumerate}
\usepackage{graphics} 
\usepackage{epsfig} 
\usepackage{mathptmx} 
\usepackage{times} 
\usepackage{amsmath} 
\usepackage{dutchcal}
\usepackage{makecell}

\usepackage{amssymb}  
\usepackage{mathtools}
\usepackage{bbm}
\usepackage{mathrsfs}

\usepackage{algpseudocode}
\usepackage[linesnumbered, ruled]{algorithm2e}
\SetKwRepeat{Do}{do}{while}%
\usepackage{algpseudocode}
\usepackage{listings}
\usepackage{hyperref} 
\usepackage{booktabs} 
\usepackage{xcolor} 
\usepackage{graphicx}
\usepackage{subfigure}
\usepackage{bbm}
\usepackage{enumitem}
\usepackage{amsmath,amssymb,amsthm}
\usepackage[export]{adjustbox}
\usepackage{changes}
\usepackage{multirow}

\definechangesauthor[color=purple, name=Haoyuan Sun]{SHY}

\title{A Method on Searching Better Activation Functions}

\author{%
  Haoyuan Sun$^{* , 1}$,
  Zihao Wu$^{* , 2}$,
  \textbf{Bo Xia}$^{1}$,
  \textbf{Pu Chang}$^{3}$,
  \textbf{Zibin Dong}$^{2}$,
  \textbf{Yifu Yuan}$^{2}$,
  \vspace{3pt}
  \\
  \textbf{Yongzhe Chang}$^{\dagger,1}$, 
  \textbf{Xueqian Wang}$^{\dagger, 1}$
  \vspace{3pt}
  \\
  $^{*}$equal contribution 
  \vspace{.1em} \hspace{4pt}
  $^{\dagger}$corresponding authors
  \vspace{3pt}
  \\
  $^{1}$Tsinghua University \vspace{.1em} \hspace{4pt}
  $^{2}$Tianjin University \vspace{.1em} \hspace{4pt}
  $^{3}$Anhui Polytechnic University
  \vspace{3pt}
}

\begin{document}

\maketitle

\begin{abstract}
The success of artificial neural networks (ANNs) hinges greatly on the judicious selection of an activation function, introducing non-linearity into network and enabling them to model sophisticated relationships in data. However, the search of activation functions has largely relied on empirical knowledge in the past, lacking theoretical guidance, which has hindered the identification of more effective activation functions. In this work, we offer a proper solution to such issue. Firstly, we theoretically demonstrate the existence of the worst activation function with boundary conditions (WAFBC) from the perspective of information entropy. Furthermore, inspired by the Taylor expansion form of information entropy functional, we propose the Entropy-based Activation Function Optimization (EAFO) methodology. EAFO methodology presents a novel perspective for designing static activation functions in deep neural networks and the potential of dynamically optimizing activation during iterative training. Utilizing EAFO methodology, we derive a novel activation function from ReLU, known as Correction Regularized ReLU (CRReLU). Experiments conducted with vision transformer and its variants on CIFAR-10, CIFAR-100 and ImageNet-1K datasets demonstrate the superiority of CRReLU over  existing corrections of ReLU. Extensive empirical studies on task of large language model (LLM) fine-tuning, CRReLU exhibits superior performance compared to GELU, suggesting its broader potential for practical applications.
\end{abstract}

\section{Introduction}

Flourishing development of artificial intelligence is predominantly attributable to rapid advancements in artificial neural networks (ANNs) observed in recent years. Activation functions (AFs) play a critical role in the performance of ANNs due to their fundamental role in enabling nonlinear representations. Despite continuous development of novel activation functions and their empirical success in improving network performance, theoretical analysis towards these activation functions remain scarce in the research literature. In other words, proposal of improved activation functions is often based on empirical evidence without theoretical validations, which greatly hinders the search for better activation functions. Hence, a theoretically reliable methodology on searching better activation functions holds significant value for the machine learning community and future research.

In our work, we initiate our exploration from the correlation between information entropy and Bayesian error rate. Subsequently, we establish the connection between activation function and information entropy, ultimately deriving the specific form that the worst activation function does exist under boundary conditions. Based on the derivation, we propose a novel method for optimizing activation functions, namely \textbf{the Entropy-based Activation Function Optimization(EAFO)} methodology. Utilizing EAFO methodology, we derive a novel activation function known as \textbf{Correction Regularized ReLU (CRReLU)} with the beginning of conventional ReLU \citep{article, 5459469, Nair2010RectifiedLU}. The derived CRReLU activation function possesses several desirable properties, including the avoidance of neuron death, the preservation of neuron sparsity, and so on. Experiments involving the vision transformer \citep{dosovitskiy2020image} and its variants \citep{pmlr-v139-touvron21a,han2021transformer}, conducted on CIFAR-10, CIFAR-100 \citep{krizhevsky2009learning} and ImageNet-1K \citep{5206848} datasets, have consistently demonstrated the superior performance of CRReLU compared to other activation function baselines. Extensive experimental studies on the task of large language model (LLM) fine-tuning with direct preference optimization (DPO) method \citep{rafailov2023direct} also demonstrate that CRReLU surpasses GELU in performance, suggesting the wider applicability of CRReLU in practical scenarios. Moreover, the EAFO methodology also shows potential to further optimize activation functions during the iterative training of ANNs, although the specific optimization techniques remain a topic of ongoing research.

In summary, our main contributions are as follows:
\begin{itemize}
\item We theoretically prove the existence of the worst activation function with boundary conditions from the perspective of information entropy; and starting from the worst activation function, performance of activation functions always improves.
\item We propose the Entropy-based Activation Function Optimization (EAFO) methodology, which provides a novel perspective for designing static activation functions in deep neural networks and the potential of dynamically optimizing activation during iterative training.
\item We derive a novel activation function known as Correction Regularized ReLU (CRReLU) starting from ReLU 
utilizing the EAFO methodology. Experiments across several mainstream architectures, datasets and tasks demonstrate that the proposed CRReLU exceeds existing activation functions, exhibiting exceptional performance.
\end{itemize}

\section{Related Work}
\label{related work}
With the development of deep learning, deep neural networks (DNNs) have gained significant prominence and achieved notable success across various domains. Recent advancements in the field of natural language processing, exemplified by large language models such as GPT-4 \citep{openai2023gpt4}, LLama-3 \citep{touvron2023llama}, and Gemini \citep{gemmateam2024gemma}, have propelled machine understanding and generation of natural language to unprecedented levels of accuracy. Additionally, deep neural networks have also achieved important applications in computer vision \citep{dosovitskiy2020image,pmlr-v139-touvron21a,han2021transformer}, deep reinforcement learning \citep{schulman2017proximal}, autonomous driving\citep{pan2024vlp}, and many other areas.

The nonlinearity of activation functions in neural networks is crucial for both enabling the efficient learning of complex patterns, and facilitating the extraction of intricate and hierarchical representations from input data, thus allowing them to capture more complex relationships between input and output variables. In contrast, however, the nonlinear activation functions of deep neural networks also presents challenges during training, encompassing challenges like gradient vanishing \citep{279181}, gradient exploding \citep{Larochelle2009ExploringSF}, and so on.

To address these challenges, researchers have explored alternative approaches for improvement, including the enhancement of activation functions. In the nascent stages of activation function development, scholars predominantly focused on rudimentary thresholding functions, initially directing their attention towards squashing functions such as the Sigmoid function and the Tanh function \citep{Hornik1991ApproximationCO}. In order to mitigate the issues of vanishing and exploding gradients, various non-squashing functions have been proposed. Notably, ReLU \citep{article,5459469,Nair2010RectifiedLU} has played a pivotal role in the remarkable success of deep learning. The derivative of ReLU for positive inputs is one, thereby preventing the gradient from vanishing; however, negative values are mapped to zero, leading to two main issues: (1) The absence of information flow for negative values, known as dying ReLU ; (2) The shift in subsequent layers due to positive bias maintained by activation.

Given the aforementioned challenges, researchers have dedicated significant efforts to improving the effectiveness of activation functions. The Leaky ReLU \citep{Maas2013RectifierNI} activation function permits a small negative slope, ensuring some gradient can still be  propagated even when input is less than zero. The Parametric ReLU (PReLU) \citep{he2015delving} is an extension of the Leaky ReLU, where $\alpha$ is considered a learnable parameter that is learned from data rather than being predetermined. The Exponential Linear Unit (ELU) \citep{clevert2016fast} outputs a negative value when $x$ is less than 0, leading to the advantageous property of the average output approaching 0. The Continuously Differentiable Exponential Linear Unit (CELU) \citep{barron2017continuously} proposes an alternative parameterization that simplifies analysis of the rectifier function and facilitates the tuning process of parameters in ELU. The Swish (also known as SiLU) \citep{2017Searching} has been shown to enhance training stability and performance in deep learning models due to its smooth nature and improved gradient propagation. In Mish \citep{2020Mish} activation function, unboundedness of positive values avoids the saturation led by a plateau, slight allowance for negative values enables better gradient flow, and the smoother activation function allows better information to flow deep into neural networks, thus resulting in better accuracy and generalization in performance.

\section{Motivation}
\label{motivaiton}
In Section \ref{related work}, it is apparent that researchers have dedicated substantial efforts to the exploration of improved activation functions, which are widely acknowledged to hold considerable significance for the advancement of deep learning. However, it has also come to our attention that proposals for these activation functions lack a theoretical framework, indicating such searches are, to some extent, inefficient and aimless.

GELU(Gaussian Error Linear Unit)\citep{hendrycks2023gaussian} was first proposed in 2016 and has since gained significant success in a variety of fields, especially with the emergence of large language models in recent years. It has been successfully incorporated into several cutting-edge neural network architectures, such as BERT\citep{devlin2019bert} , ViT \citep{dosovitskiy2020image} , GPT-4\citep{openai2023gpt4} , and so on, demonstrating its versatility and effectiveness. In the work conducted by \citet{lee2023gelu} (2023), insightful mathematical properties of the GELU are finally unveiled, including its differentiability, boundedness, stationarity, and smoothness. Hence, it is often the case that superior performance exhibited by novel activation functions frequently lacks mathematical explanations for their observed enhancements. Understanding may merely limited to the fact that it exhibits improved performance, which hampers exploration for better activation functions and interpretability of neural networks.

In light of the aforementioned challenges, our work endeavors to propose a methodology for searching better activation functions, not only enabling the discovery of improved activation functions but also elucidating the reasons behind their superior performance at the same time.

\section{Methodology}
\label{Methodology}
\subsection{Problem Setup}
\subsubsection{Bayesian Error Rate and Information Entropy}
A deep neural network can be simplified as comprising a feature extraction layer, which is subsequently followed by a fully connected layer for final classification. From a probabilistic perspective, in binary classification, the feature extraction layer can be conceptualized as transforming the shape of mixture distribution, thereby enabling the final fully connected layer to separate two distributions with a hyperplane. Hence, the more overlapping two distributions are, the higher Bayesian error rate and the worse classification performance. Furthermore, a lower information entropy corresponds to a higher likelihood of forming two distinct peaks (i.e. the smaller classification uncertainty, the easier to classify); and an increase in the overlap between two distributions also leads to the increase of information entropy (i.e. the greater classification uncertainty, the harder to classify). In addition, the above statements can be extended to multi-class classification, and further elaboration is omitted here.
\subsubsection{Activation Function and Information Entropy}
\label{Activation Function and Information Entropy}
Assuming the inverse function of the activation function is $y (x)$, and the activation function is monotonically increasing. Many previous activation functions, such as Sigmoid and Tanh \citep{Hornik1991ApproximationCO}, satisfy the assumption that the function has an inverse function in entire definition domain. Furthermore, when an activation function fails to meet the assumption, we can transform the part of such function satisfying this assumption, as is the case with the positive part of ReLU.

Then we set data distribution before passing through the activation function obeys the distribution $p(x)$. Thus, data distribution after passing through activation function is : $q(x)=p\left(y(x)\right)y'(x)$, where $y'(x)$ represents the derivative of $y(x)$. Hence, we can express the information entropy as:
\[
\mathbb{H}(y(x)) = - \int q(x) \log q(x) \text dx
     = - \int p(y(x))y'(x) \log(p(y(x))y'(x))\text dx
    =\int \mathbb{G}(y '(x),y (x))\text dx
\]
Therefore, the information entropy can be deemed as a \textbf{functional}, which takes a function $y(x)$ as input and produces a real number as output.

\subsection{Worst Activation Function with Boundary Condition (WAFBC)}
\label{Worst Activation Function with Boundary Condition (WAFBC)}
Firstly, we would like to determine the extremum (whether it is a maximum or minimum) of the functional $\mathbb{H}(y(x))$. For further deductions, taking the simplest functional into consideration, e.g. setting $\mathbb{H}(y(x))= \int \mathbb{G}\left(y'(x),y(x),x\right) \text dx$.

In order to research the influence brought by variations of function $y(x)$, we apply a small perturbation $\varepsilon\eta(x)$ to function $y(x)$, and then the functional $\mathbb{H}\left(y(x)+\varepsilon\eta(x)\right)$ takes the form as:
\[
\mathbb{H}\left(y(x)+\varepsilon\eta(x)\right)=\int \mathbb{G}(y'(x)+\varepsilon\eta'(x),y(x)+\varepsilon\eta(x),x)\text dx
\]
We apply \textbf{Taylor expansion} to functional $\mathbb{H}\left(y(x)+\varepsilon\eta(x)\right)$, we can obtain the following equation:
\begin{equation}
\label{Taylor expansion}
\begin{split}    
&\mathbb{H}\left(y(x)+\varepsilon\eta(x)\right)
=\int \left[\mathbb{G}(y'(x),y(x),x)+\varepsilon\frac{\partial \mathbb{G}}{\partial y'}\eta'(x)+\varepsilon\frac{\partial \mathbb{G}}{ \partial y}\eta(x)+\mathcal{O}(\varepsilon)\right]\text dx\\
=&\mathbb{H}(y(x))+\varepsilon\int \left[\frac{\partial \mathbb{G}}{\partial y}\eta(x)+\frac{ \partial \mathbb{G}}{\partial y'}\eta'(x)\right]\text dx+\mathcal{O}(\varepsilon)
\end{split}
\end{equation}

As illustrated in Section \ref{Activation Function and Information Entropy}, $q(x)=p\left(y(x)\right)y'(x)$ is the data distribution  after passing through activation function. We can easily get that for the inverse function $y(x)$ of activation function, when $x$ approaches the lower bound (e.g. the initial activation function value approaches lower bound), $y(x)$ should approaches negative infinity; and when $x$ approaches the upper bound (e.g. the initial activation function value approaches upper bound), $y(x)$ should approaches positive infinity. And for the sake that $\varepsilon\eta(x)$ is a small perturbation applied to $y(x)$, we can draw the conclusion that $\eta(x)$ must be 0 at the boundaries. 

Utilizing the method of integration by parts and boundary condition towards Equation \ref{Taylor expansion}, we can derive the following results:
\[
\int\frac{ \partial \mathbb{G}}{\partial y'}\eta'(x)\text dx=\int\frac{\partial \mathbb{G}}{\partial y'}\text d\eta(x)
=\eta(x)\frac{\partial \mathbb{G}}{\partial y'}\bigg|_{x}-\int\eta(x)\frac{\text d}{\text dx}\left(\frac{\partial \mathbb{G}}{\partial y'}\right) \text dx
=-\int\eta(x)\frac{\text d}{\text dx}\left(\frac{\partial \mathbb{G}}{\partial y'}\right) \text dx
\]
Thus, $\mathbb{H}\left(y(x)+\varepsilon\eta(x)\right)$ has the following expression:
\[
\mathbb{H}\left(y(x)+\varepsilon\eta(x)\right)=\mathbb{H}(y(x))+\varepsilon\int \left[\frac{\partial \mathbb{G}}{\partial y}-\frac{\text d}{\text dx}\left(\frac{\partial \mathbb{G}}{\partial y'}\right)\right]\eta(x)\text dx+\mathcal{O}(\varepsilon)
\]
In analogy to the extremum of ordinary functions, it is expected that the first-order term should be 0 at the extremum point. Such requirement for arbitrary $\eta(x)$ leads to the Euler-Lagrange equation:

\begin{equation}
\label{EL equation1}
\frac{\text d}{\text dx}(\frac{\partial \mathbb{G}}{\partial y'})-\frac{\partial \mathbb{G}}{ \partial y}=0
\end{equation}

\begin{proposition}
\label{ Euler-Lagrange equation}
If $\mathbb{G}$ is independent of $x$, i.e. $\mathbb{G}=\mathbb{G}(y,y')$, based on the Euler-Lagrange equation expressed in Equation \ref{EL equation1}, then we have:

\begin{equation}
\label{Euler-Lagrange equation_equation}
\mathbb{G} - y' \frac{\partial\mathbb{G}}{\partial y'}=C
\end{equation}

\end{proposition}

Detailed proof of Proposition \ref{ Euler-Lagrange equation} can be seen in Appendix \ref{Proof of Euler-Lagrange equation}.

Substitute $\mathbb{G}=p(y(x))y'(x) \log(p(y(x))y'(x))$ into Equation \ref{Euler-Lagrange equation_equation} and perform the calculation, the final result is:
\[
\frac{\text dy}{\text dx}p(y(x))=C
\]
Integrating both sides of the equation simultaneously, the final solution is:

\begin{equation}
\label{solution}
    x=c_1\int p(y)\text d y+c_2
\end{equation}
Based on the solution we get in Equation \ref{solution}, for the sake that $y(x)$ is the inverse function of the activation function, the first integral equation can finally be solved to obtain the form of the activation function as:
\begin{equation}
\label{WAFBC}
f(x)=C_1{\int_{-\infty}^x p(t)\text dt}+C_2
\end{equation}
, where $C_1$ and $C_2$ are two constants based on the upper bound and lower bound of activation function.

Equation \ref{WAFBC} shows the analytical form of the worst activation function with boundary condition. We provide further discussion on this form in Appendix \ref{Further Discussion on WAFBC}. Through the above derivation, extremum of the functional is determined. Furthermore, we would like to deduce whether it is a maximum value or a minimum one. Applying Legendre condition to the functional extremum, then we have:
\[
\mathbb{G}_{y^{\prime}y^{\prime}}=-\frac{p(y(x))}{y'}\leqslant 0
\]
Therefore, the derived extremum is a maximum extremum, and is a global maximum extremum actually, meaning the deduced activation function has the worst performance. Actually, the WAFBC possesses some intriguing properties, for example, it inherently has upper and lower bounds, which can explain why bounded activation functions like Sigmoid and Tanh do not perform as well as unbounded functions like ReLU.

\subsection{Entropy-based Activation Function Optimization (EAFO)}
\label{Entropy-based Activation Function Optimization (EAFO)}
In Section \ref{Worst Activation Function with Boundary Condition (WAFBC)}, we have derived the extremum of the functional, showing the analytical form in Equation \ref{WAFBC}. However, the solution obtained is the global maximum, rather than the minimum. The minimum of the functional is needed if we would like to obtain the best activation function. Nonetheless, based on calculation, the actual situation is that this functional only has a global maximum but no global minimum exists. \textbf{Hence, there is no best activation function, but only better activation functions.} In this scenario, WAFBC represents a global maximum of the functional, indicating that the performance of activation functions consistently improves from WAFBC to any alternative activation functions. Therefore, we propose the following question: Is there a methodology to begin with an existing, high-performing activation function, and subsequently develop an activation function with superior performance?

Let's reconsider the Taylor expansion of the functional \[
\mathbb{H}(y(x)+\varepsilon\eta(x))=\mathbb{H}(y(x))+\varepsilon\int \left[\frac{\partial \mathbb{G}}{\partial y}-\frac{\text d}{\text dx}\left(\frac{\partial \mathbb{G}}{\partial y'}\right)\right]\eta(x)\text dx+\mathcal{O}(\varepsilon)\]
To minimize the information entropy of novel activation function, it is advisable to reduce the first-order term of Taylor expansion. In order to ensure that the information entropy of novel activation function has been indeed reduced, we would like to set $\eta(x)$ as the opposite sign to $\frac{\partial \mathbb{G}}{\partial y}-\frac{\text d}{\text dx}\left(\frac{\partial \mathbb{G}}{\partial y'}\right)$, which means we set:
\begin{equation}
\label{etax}
    \eta(x)=-\left(\frac{\partial \mathbb{G}}{\partial y}-\frac{\text d}{\text dx}\left(\frac{\partial \mathbb{G}}{\partial y'}\right)\right)
\end{equation}
Substitute the analytical form of functional $\mathbb{G}(y '(x),y (x))$ into Equation \ref{etax}, perform the calculation, we can derive the following equation:
\begin{equation}
\label{etax_done}
    \eta(x)=-\left(p(y(x))\frac{y''(x)}{y'(x)}+p'(y(x))y'(x)\right) 
\end{equation}
, where $p(x)$ is the probability density function (PDF) of data distribution before passing through the activation function; $p'(x)$ is the first order derivative of PDF; $y(x)$ is inverse function of the activation function; $y'(x)$ is  the first order derivative of $y(x)$; $y''(x)$ is the second order derivative of $y(x)$.

As a result, we have derived a correction term that is capable of decreasing information entropy, expressing its general form in Equation \ref{etax_done}. Subsequently, we can obtain the inverse function of the optimized activation function, denoted as $g(x) = y(x) + \eta(x)$. Finally, the optimized activation function can be obtained by deriving the inverse function of $g(x)$.

\textbf{EAFO methodology outline }. In summary, we express the theoretical EAFO methodology as follows: 1) Utilize Equation \ref{etax_done} and derive correction term $\eta(x)$ given data distribution $p(y)$ and inverse function of activation function $y(x)$. 2) Sum the correction term with the inverse function to obtain the inverse function of the optimized function, i.e. $g(x) = y(x) + \eta(x)$ . 3) Derive the rigorous or approximate inverse function of $g(x)$, yielding the optimized activation function.

Furthermore, EAFO methodology has also shown the potential of dynamically optimizing activation during iterative training. We are acknowledged that activation of neural networks with Multi-Layer Perceptrons (MLPs) architecture is typically fixed. Recent studies, such as work done by \citet{liu2024kan}, have suggested the optimization of activation in innovative network architectures (Kolmogorov-Arnold Networks). 
Furthermore, across true data distributions $p(y)$, utilizing EAFO methodology, we may continuously optimize activation $y(x)$ practically under Multi-Layer Perceptrons (MLPs) architecture with numerical methods. Moreover, in theory, it is feasible to optimize activation functions using methods such as gradient descent optimization of the information entropy
functional through numerical methods; however, we are also aware that this would result in an explosion of
computational complexity in large neural networks, which calls for practically efficient
algorithms. Hence, the EAFO methodology is still in the theoretical stage presently, providing guidance for calculating the analytical form of better
activation functions.

\subsection{Correction Regularized ReLU (CRReLU) : From ReLU to Better}
As illustrated in Section \ref{Worst Activation Function with Boundary Condition (WAFBC)}, it is theoretically true that the worst activation function exists, and we can determine its exact form. Actually, beginning with the worst activation function, the value of the functional $\mathbb{G}$ consistently decreases, indicating an improvement in the performance of activation function. This reveals the feasibility of searching an improved activation function, which constitutes the crux of "optimization". In Section \ref{Entropy-based Activation Function Optimization (EAFO)}, EAFO is proposed as the optimization methodology. Hence, we can easily think of optimizing from WAFBC to get a better-performing activation function. While it is true that such an idea is feasible, we also observe that WAFBC itself takes the form of a variable upper bound integral, which yields a complex form of $\eta(x)$ and renders the deduced result not practically significant. Moreover, commencing optimization from WAFBC also leads to sluggish advancement. Therefore, in practical applications, we are inclined to start from an activation function that already demonstrates relatively good performance.

Here, we would like to take ReLU \citep{article, 5459469, Nair2010RectifiedLU} as the beginning, and show the process of finding a better activation function. Before the deduction, we also notice that ReLU is lack of an inverse function over the entire domain. In this section, we would like to utilize following strategies for mitigating the aforementioned dilemma: the initial activation function only necessitates an inverse function in specific regions where it is required; and when encountering parts without an inverse function, we may employ practical approximations. Therefore, we initially examine the region where $x$ is positive in the case of ReLU. As shown in Equation \ref{etax_done}, the derivation of correction term $\eta(x)$ only requires original distribution $p(y)$ and inverse function of the activation function $y(x)$. Knowledge of activation function is easily available, whereas original distribution remains unexplored. However, in real experiments, original distribution of experimental data would surely exhibit a substantial degree of morphological variability, thus lacking a perfect analytical form. Hence, we assume the situation is that networks are large enough, according to the Central Limit Theorem, the data processed by them can be approximated as a Gaussian distribution \citep{Williams1996ComputingWI,lee2018deep,park2020nngpguided,gao2023wide}\citep{huang2021convolutionweightdistribution}. Certainly, such assumption may not always hold in networks of real experiments; nevertheless, approximation of the exact solution for inverse function and existence of the learnable parameter $\epsilon$ have significantly mitigated the impact of such assumption, which can also be demonstrated by the insensitivity of CRReLU to data distribution shown in Section \ref{experiments}.

Now, let's consider the derivation from ReLU to CRReLU. For the sake of concise representation, we rewrite the data distribution and the derivative of data distribution as: \[
p(y)=C\cdot e^{-\frac{y^2}{2}} \quad,\quad
 p'(y)=-C\cdot ye^{-\frac{y^2}{2}}
\]

Furthermore, ReLU has a mathematical function defined as $y = x$ when $x$ is positive, meaning we have $y(x)=x$ , $y'(x)=1$ and $y''(x)=0$. Therefore,
\[
p'(y(x)) = p'(x) = -C\cdot ye^{-\frac{y^2}{2}} = -C\cdot xe^{-\frac{x^2}{2}}
\]
Ultimately, by incorporating $p'(y) = -C\cdot xe^{-\frac{x^2}{2}}$ , $y'(x)=1$ and $y''(x)=0$ into Equation \ref{etax_done}, we can obtain:
\[
\eta(x) = -C \cdot xe^{-\frac{x^2}{2}}
\]

 Furthermore, we make constant $C$ as a learnable parameter $\epsilon$ with the purpose of enabling self-optimization in networks. According to EAFO methodology, we can get the inverse function of revised activation function as follows: 
\begin{equation}
\label{CRReLU before inverse}
    g(x) = x - \epsilon x e^{-\frac{x^2}{2}} \quad \quad x \geqslant 0
\end{equation}

Finally, the optimized activation function CRReLU can be obtained by deriving the inverse function of $g(x)$. However, obtaining the inverse function of Equation \ref{CRReLU before inverse} presents a challenge using conventional methods; as a consequence, we use the following function as a form of practical approximation.
\begin{equation}
\label{CRReLU after inverse}
    f(x) = x + \epsilon x e^{-\frac{x^2}{2}} \quad \quad x \geqslant 0
\end{equation}

We show the rationalization and reliability of utilizing Equation \ref{CRReLU after inverse} as the approximate inverse function of Equation \ref{CRReLU before inverse} in Proposition \ref{proposition for CRReLU's inverse function}

\begin{proposition}
\label{proposition for CRReLU's inverse function}
Known $g(x)=x-\epsilon xe^{-\frac{x^2}{2}}, f(x)=x+\epsilon xe^{-\frac{x^2}{2}}$, for $x\geqslant0$ , the absolute value of error between $g\left(f(x)\right)$ and $x$ is bounded with $\left|e^{-1}\epsilon^2 + 0.5e^{-\frac{3}{2}}\epsilon^3\right|$.
\end{proposition}
Detailed proof of Proposition \ref{proposition for CRReLU's inverse function} can be seen in Appendix \ref{proof of proposition for CRReLU's inverse function}.

 As illustrated in Section \ref{Worst Activation Function with Boundary Condition (WAFBC)}, $\epsilon \eta(x)$ is the small perturbation; hence, from a theoretical perspective, we can set $\epsilon \eta(x)$ as an infinitesimal. Furthermore, in this case, given the knowledge that $\eta(x)$ is a bounded function, we can easily deduce that $\epsilon$ is also an infinitesimal. Therefore, the absolute value of error between $g\left(f(x)\right)$ and $x$ is an infinitesimal of higher order. In practice, we typically initialize $\epsilon$ to a small value, such as 0.01 (as described in Section \ref{experiments}), implying that the absolute value of error is a small value.

Finally, let's consider the part where $x$ is negative. When $x$ is negative, the inverse function of ReLU can be visualized as a ray emanating from the origin and extending to infinity, possessing an infinite slope; and when $x$ is positive, it constitutes a ray with the slope of 1. Hence, the correction term solution for both positive and negative values of $x$ can be considered identical, differing only by constant $C$. In Equation \ref{CRReLU after inverse} and Proposition \ref{proposition for CRReLU's inverse function}, it is shown that incorporating the correction term into a linear activation function can have beneficial effects by reducing the information entropy. Therefore, we can obtain the full form of Correction Regularized ReLU as:
\begin{equation}
\label{CRReLU_final}
f(x)=\text{max} (0,x)+ \varepsilon xe^{-\frac{x^2}{2}}
\end{equation}

\textbf{Discussion on introduced learnable parameter $\epsilon$.}  In Section \ref{Worst Activation Function with Boundary Condition (WAFBC)}, we have successfully demonstrated existence of the worst activation function, and from the worst as a starting point, it always moves towards improvement, regardless of the direction taken. However, commencing from a specific activation function, like ReLU here, does not invariably result in improvement across all directions, i.e. certain optimization paths may lead to deteriorated outcomes. Therefore, from the practical perspective, we introduce learnable parameter $\epsilon$ with the aim of enabling self-optimization of networks. From another perspective, in the derivation from ReLU to CRReLU, we assume that data follows Gaussian distribution, which might not be true in real experiments. Existence of the learnable parameter $\epsilon$ also weakens this assumption to some extent.

Finally, we provide further details of CRReLU in Appendix \ref{Further details of CRReLU}, including python-like pseudocode of CRReLU  in Appendix \ref{Correction Regularized ReLU (CRReLU) Pseudocode}, and further discussion on properties of CRReLU in Appendix \ref{Further Discussion on Properties of CRReLU}.
\section{Experiments}
\label{experiments}

\textbf{Datasets.} In experiments of image classification task, we adopt three datasets, ordered as CIFAR-10 \citep{krizhevsky2009learning}, CIFAR-100 \citep{krizhevsky2009learning} and ImageNet-1K \citep{5206848} in terms of the number of classification categories. In experiments of large language model (LLM) fine-tuning task, we employ two human preference datasets: SHP \citep{ethayarajh2022understanding} and HH \citep{bai2022training}.

\textbf{Baselines.} We conduct experiments comparing the performance of CRReLU with several typical existing corrections of ReLU as illustrated in Section \ref{related work} and Section \ref{motivaiton} : PReLU \citep{he2015delving}, ELU \citep{clevert2016fast}, CELU \citep{barron2017continuously}, GELU \citep{hendrycks2023gaussian}, Swish (SiLU) \citep{2017Searching} and Mish \citep{2020Mish}. 

\textbf{Experimental hyperparameters.} For all transformer-based architectures, we directly set $\epsilon$ to 0.01 without further optimization. Detailed experimental hyperparameters are provided in Appendix \ref{Details of experimental setting}.
\subsection{Task of Image Classification}
We conduct all experiments on 4$\times$RTX3090 for 100 epochs using the AdamW optimizer with weight decay of 0.05, gradient clipping norm of 1.0, cross entropy loss function, and cosine annealing learning rate scheduler with linear warm-up.

\textbf{Experiments of ViTs on CIFAR-10 and CIFAR-100.}
Vision Transformer and its variants possess sufficiently complex structure and representational capability, garnering widespread attention from the community. Moreover, the assumption of Gaussian distribution has been theoretically proved as reasonable for sufficiently large MLPs \citep{Williams1996ComputingWI, lee2018deep, park2020nngpguided, gao2023wide} and CNNs \citep{huang2021convolutionweightdistribution}; however, the distribution of data under attention mechanism of transformers remains unexplored. Hence, we select vision transformer and its variants as our test model in order to further investigate the insensitivity of CRReLU to data distribution. Phase of experiments on CIFAR-10 and CIFAR-100 involves the selection of Vision Transformer (ViT) \citep{dosovitskiy2020image}, Data-Efficient Image Transformer (DeiT) \citep{pmlr-v139-touvron21a} and Transformer in Transformer (TNT) \citep{han2021transformer}. We report the top-one accuracy on CIFAR-10 in Table \ref{Test accuracy of experiments conducted on CIFAR-10 for 100 epochs.} and CIFAR-100 in Table \ref{Test accuracy of experiments conducted on CIFAR-100 for 100 epochs.}, demonstrating CRReLU outperforms other existing corrections of ReLU on CIFAR dataset.

\begin{table}[h]
\caption{Test accuracy of experiments conducted on CIFAR-10 for 100 epochs.}
\label{Test accuracy of experiments conducted on CIFAR-10 for 100 epochs.}
\centering
\begin{tabular}{cc|cccccc|c}
\toprule
\multicolumn{2}{c|}{Top-one Accuracy}             & GELU  & ELU   & PReLU & CELU  & SiLU  & Mish  & \begin{tabular}[c]{@{}c@{}}CRReLU\\ (ours)\end{tabular} \\ \midrule
\multicolumn{1}{c|}{CIFAR-10} & ViT-Tiny  & 0.706 & 0.669 & 0.786 & 0.669 & 0.683 & 0.687 & \textbf{0.802}                                          \\
\multicolumn{1}{c|}{CIFAR-10} & DeiT-Tiny & 0.716 & 0.671 & 0.753 & 0.671 & 0.694 & 0.695 & \textbf{0.768}                                          \\
\multicolumn{1}{c|}{CIFAR-10} & TNT-Small & 0.743 & 0.689 & 0.761 & 0.689 & 0.719 & 0.725 & \textbf{0.775}                                          \\ \bottomrule
\end{tabular}
\end{table}

\begin{table}[h]
\caption{Test accuracy of experiments conducted on CIFAR-100 for 100 epochs.}
\label{Test accuracy of experiments conducted on CIFAR-100 for 100 epochs.}
\centering
\begin{tabular}{cc|cccccc|c}
\toprule
\multicolumn{2}{c|}{Top-one Accuracy}             & GELU  & ELU   & PReLU & CELU  & SiLU  & Mish  & \begin{tabular}[c]{@{}c@{}}CRReLU\\ (ours)\end{tabular} \\ \midrule
\multicolumn{1}{c|}{CIFAR-100} & ViT-Tiny  & 0.322 & 0.287 & 0.421 & 0.287 & 0.306 & 0.297 & \textbf{0.459}                                          \\
\multicolumn{1}{c|}{CIFAR-100} & DeiT-Tiny & 0.460 & 0.400 & 0.493 & 0.400 & 0.429 & 0.429 & \textbf{0.508}                                          \\
\multicolumn{1}{c|}{CIFAR-100} & TNT-Small & 0.484 & 0.435 & 0.498 & 0.435 & 0.459 & 0.464 & \textbf{0.508}                                          \\ \bottomrule
\end{tabular}
\end{table}

\textbf{Experiments of ViTs on ImageNet-1K.} ImageNet-1K dataset poses a significant challenge to information processing capability of neural networks due to its large image size and extensive range of classification categories. Hence, phase of experiments on ImageNet-1K involves the selection of Vision Transformer (ViT) \citep{dosovitskiy2020image} and Data-Efficient Image Transformer (DeiT) \citep{pmlr-v139-touvron21a}. We report the top-one accuracy on ImageNet-1K in Table \ref{Test accuracy of experiments conducted on ImageNet-1K for 100 epochs.}. 

\begin{table}[h]
\caption{Test accuracy of experiments conducted on ImageNet-1K for 100 epochs.}
\label{Test accuracy of experiments conducted on ImageNet-1K for 100 epochs.}
\centering
\begin{tabular}{cc|cccccc|c}
\toprule
\multicolumn{2}{c|}{Top-one Accuracy}             & GELU  & ELU   & PReLU & CELU  & SiLU  & Mish  & \begin{tabular}[c]{@{}c@{}}CRReLU\\ (ours)\end{tabular} \\ \midrule
\multicolumn{1}{c|}{ImageNet-1K} & ViT-Tiny  & 0.542 & 0.384 & 0.572 & 0.384 & 0.469 & 0.479 & \textbf{0.579}                                          \\
\multicolumn{1}{c|}{ImageNet-1K} & DeiT-Tiny & \textbf{0.619} & 0.497 & 0.612 & 0.497 & 0.584 & 0.592 & \textbf{0.615}                        \\ 
\bottomrule
\end{tabular}
\end{table}
Experiments on ViT clearly demonstrate superiority of CRReLU over other activation functions, and those on DieT, GELU shows 0.4\%  higher accuracy compared to CRReLU. Such result is attributed to the teacher-student strategy structure of DieT model. We utilize the fine-tuned "deit-tiny-patch16-224" model as teacher model, which is trained with GELU. As explained in the work \citep{abnar2020transferring}, through distillation, transformers will inherit inductive bias. Hence, training a student model with GELU on ImageNet-1K with the help of teacher model, which has already been pre-trained on ImageNet-1K with GELU, is certain to achieve better results than other activation functions. 

\subsection{Task of Large Language Model (LLM) Fine-tuning}
In order to further validate the effectiveness of CRReLU on larger networks and generalization to a richer range of applications, we further perform supplementary experiments on LLM fine-tuning task. We employ the Direct Preference Optimization (DPO) \citep{rafailov2023direct} method to fine-tune GPT-2 \citep{radford2019language} on Stanford Human Preferences (SHP) dataset \citep{ethayarajh2022understanding} and Anthropic HH dataset \citep{bai2022training}. The parameter number of GPT-2 is 137 M, a relatively modest magnitude, hence we conduct full fine-tuning instead of LoRA-based one on 2$\times$RTX3090. Firstly, we carry out supervised fine-tuning (SFT) with the purpose of mitigating distribution shift between the true reference distribution which is unavailable, and the reference policy utilized by DPO. Subsequently, we separately set the penalty coefficient $\beta$ as \textbf{0.1, 1, 2, and 5}, in order to compare the performance of CRReLU and GELU under different penalty coefficients, and then execute DPO. We report evaluation metrics of fine-tuning process in Table \ref{Metrics comparison between CRReLU and GELU in the task of LLM fine-tuning.}, demonstrating CRReLU generally outperforms GELU in LLM fine-tuning task.
\begin{table}[h]
\caption{Metrics comparison between CRReLU and GELU in the task of LLM fine-tuning.}
\label{Metrics comparison between CRReLU and GELU in the task of LLM fine-tuning.}
\centering
\begin{tabular}{cc|ccc}
\toprule
\multicolumn{2}{c|}{Evaluation Metrics}&Evaluation Margin Reward$\uparrow$&Evaluation Accuracy$\uparrow$&Evaluation Loss$\downarrow$ \\ 
\midrule
\multicolumn{1}{c|}{\multirow{2}{*}{\small$\beta$ = 0.1}} & CRReLU & \textbf{0.1428}& \textbf{0.6210}& \textbf{0.6476} \\
\multicolumn{1}{c|}{}& GELU& 0.1419& 0.6196& 0.6480          \\ \midrule
\multicolumn{1}{c|}{\multirow{2}{*}{\small$\beta$ = 1}}   & CRReLU & \textbf{0.4626}          & \textbf{0.5756}     & \textbf{0.9201} \\
\multicolumn{1}{c|}{}                     & GELU   & 0.4556                   & 0.5731              & 0.9298          \\ \midrule
\multicolumn{1}{c|}{\multirow{2}{*}{\small$\beta$ = 2}}   & CRReLU & \textbf{0.7736}          & \textbf{0.5628}     & \textbf{1.462}  \\
\multicolumn{1}{c|}{}                     & GELU   & 0.7176                   & 0.5606              & 1.481           \\ \midrule
\multicolumn{1}{c|}{\multirow{2}{*}{\small$\beta$ = 5}}   & CRReLU & \textbf{1.846}           & \textbf{0.5635}     & \textbf{3.268}  \\
\multicolumn{1}{c|}{}                     & GELU   & 1.651                    & 0.5566              & 3.305           \\ \bottomrule
\end{tabular}
\end{table}

\section{Discussion}
\label{Discussion}
Pursuit of better activation functions has been a longstanding and fundamental topic in the realm of machine learning. However, prior research has consistently concentrated on empirical search, without an emphasis on understanding the underlying mathematical mechanisms. This work aims to offer a proper solution to such issue. Our investigation into the relationship between activation functions and information theory concepts reveals that information entropy can be represented as a functional. Existence of the worst activation function with boundary condition (WAFBC) furnishes a solid theoretical basis for exploring better activation functions. In the process of solving WAFBC, we draw inspiration from the Taylor expansion form, leading us to propose Entropy-based Activation
Function Optimization (EAFO) methodology. EAFO methodology presents a novel perspective for designing static activation functions in deep neural networks and shows the potential of dynamically optimizing activation during iterative training. Utilizing EAFO methodology, we derive a novel activation function from ReLU, called Correction Regularized ReLU (CRReLU). Experiments involving image classification task and large language model (LLM) fine-tuning task demonstrate that CRReLU is comparable to or surpasses existing corrections of ReLU. Overall, the EAFO methodology provides numerous promising avenues for future research on activation functions, and the CRReLU introduces a novel addition to the set of high-performing activation functions. 

\textbf{Limitations and Future Work.} Our findings raise several important questions for future work. \textit{Firstly}, how can EAFO framework be systematically generalized to non-invertible activation functions? In the initial setting of EAFO methodology, the choice of activation function is restricted to those with invertible counterparts. Despite ReLU being a prominent example of activation function without an inverse, we derive CRReLU utilizing EAFO; however, the derivation also partly benefits from the simplicity of ReLU's form and several heuristic approaches. \textit{Secondly}, how to effectively implement activation function iteration optimization during neural network training? Notwithstanding the demonstrated feasibility of iterative activation function optimization during neural network training, it is currently hindered by the high computational complexity, particularly in large-scale neural networks. Applicability of the EAFO methodology to optimize activation in alternative network structures, such as Kolmogorov-Arnold Networks (KANs), also deserves further in-depth research. Therefore, the development of practical and efficient algorithms is an exciting direction for future work. \textit{Finally}, while we have empirically validated the exceptional performance of CRReLU on image classification task and large language model fine-tuning task, its performance on other tasks remains to be explored, thereby warranting further investigation.


\newpage
\bibliographystyle{unsrtnat}  
\bibliography{reference}


\appendix
\newpage
\section{Proof of Proposition \ref{ Euler-Lagrange equation}}
\label{Proof of Euler-Lagrange equation}
\begin{proof}
    From Equation \ref{EL equation1}, we know that:
    \[
    \frac{\text d}{\text dx}(\frac{\partial \mathbb{G}}{\partial y'})-\frac{\partial \mathbb{G}}{ \partial y}=0
    \]
    Considering the total differential of $\mathbb{G}$:
    \[
    \frac{\text d\mathbb{G}}{\text dx}\left(y' , y ,x \right)=\frac{\partial \mathbb{G}}{\partial x}\cdot \frac{\text dx}{\text dx}+\frac{\partial \mathbb{G}}{\partial y}\cdot \frac{\text dy}{\text dx}+\frac{\partial \mathbb{G}}{\partial y'}\cdot \frac{\text dy'}{\text dx}=\frac{\partial \mathbb{G}}{\partial x}+\frac{\partial \mathbb{G}}{\partial y}\cdot y' + \frac{\partial \mathbb{G}}{\partial y'}\cdot y''
    \]
    Thus, we have:
    \[
    \begin{split}
    \frac{\text d}{\text dx}\left(y'\frac{\partial \mathbb{G}}{\partial y'} \right)&=y''\frac{\partial \mathbb{G}}{\partial y'}+y'\frac{\text d}{\text dx}\left(\frac{\partial \mathbb{G}}{\partial y'}\right)\\
    &=\frac{\text d\mathbb{G}}{\text dx}\left(y' , y ,x \right)-\frac{\partial \mathbb{G}}{\partial y}\cdot y'-\frac{\partial \mathbb{G}}{\partial x} + y'\frac{\text d}{\text dx}\left(\frac{\partial \mathbb{G}}{\partial y'}\right)\\
    &=\frac{\text d}{\text dx} \mathbb{G} \left(y' , y ,x \right)-\frac{\partial \mathbb{G}}{\partial x}-y'\cdot \left( \frac{\partial \mathbb{G}}{\partial y} - \frac{\text d}{\text dx}\left(\frac{\partial \mathbb{G}}{\partial y'}\right) \right)\\
    &=\frac{\text d}{\text dx} \mathbb{G} \left(y' , y ,x \right)-\frac{\partial \mathbb{G}}{\partial x}
    \end{split}
    \]
    Therefore, we know that
    \[
    \frac{\partial \mathbb{G}}{\partial x}-\frac{\text d}{\text dx}\left(\mathbb{G}-y'\frac{\partial \mathbb{G}}{\partial y'}\right)=0
    \]
    For the sake that $\mathbb{G}$ is independent of $x$, then we have that $\frac{\partial \mathbb{G}}{\partial x}=0$.
    Hence, 
    \[
    \frac{\text d}{\text dx}\left(\mathbb{G}-y'\frac{\partial \mathbb{G}}{\partial y'}\right)=0
    \]
    Finally, we can draw the conclusion that:
    \[
    \mathbb{G}-y'\frac{\partial \mathbb{G}}{\partial y'}=C
    \], which completes the proof.
\end{proof}
\section{Further Discussion on WAFBC}
\label{Further Discussion on WAFBC}
\begin{figure}[ht]
    \includegraphics[width=1\linewidth]{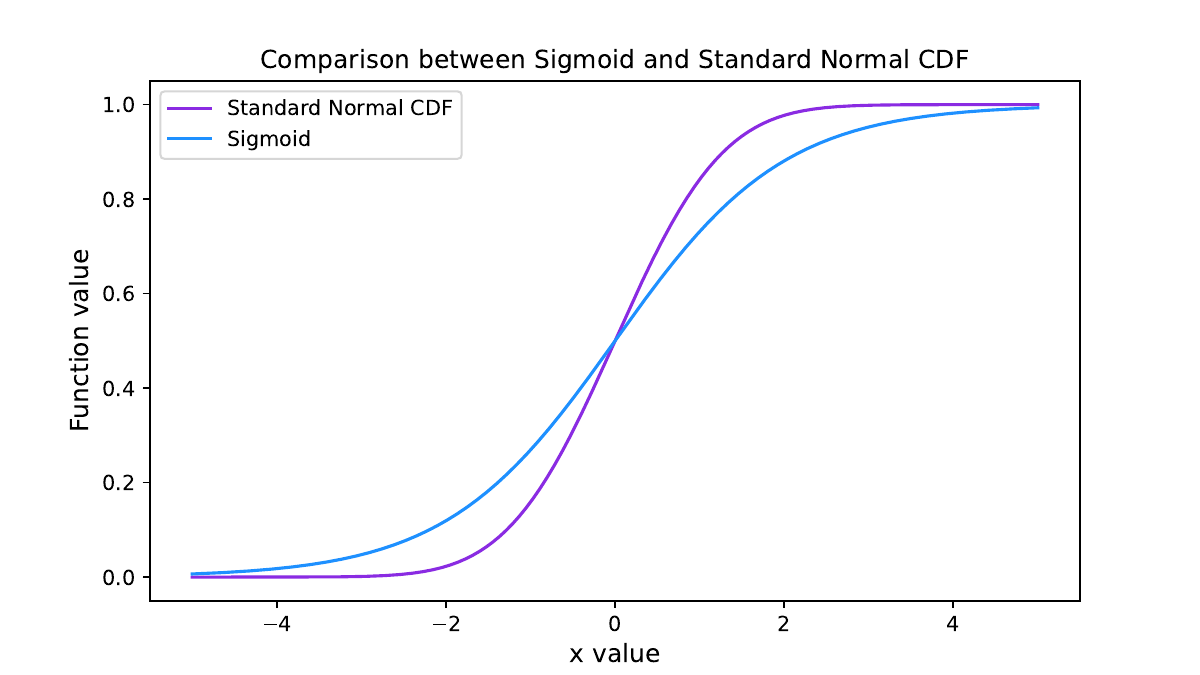}
    \vspace{-2em}
    \caption{Comparison between Sigmoid and standard normal CDF}
    \label{Comparison between Sigmoid and Standard Normal CDF}
\end{figure}

\begin{figure}[ht]
    \includegraphics[width=1\linewidth]{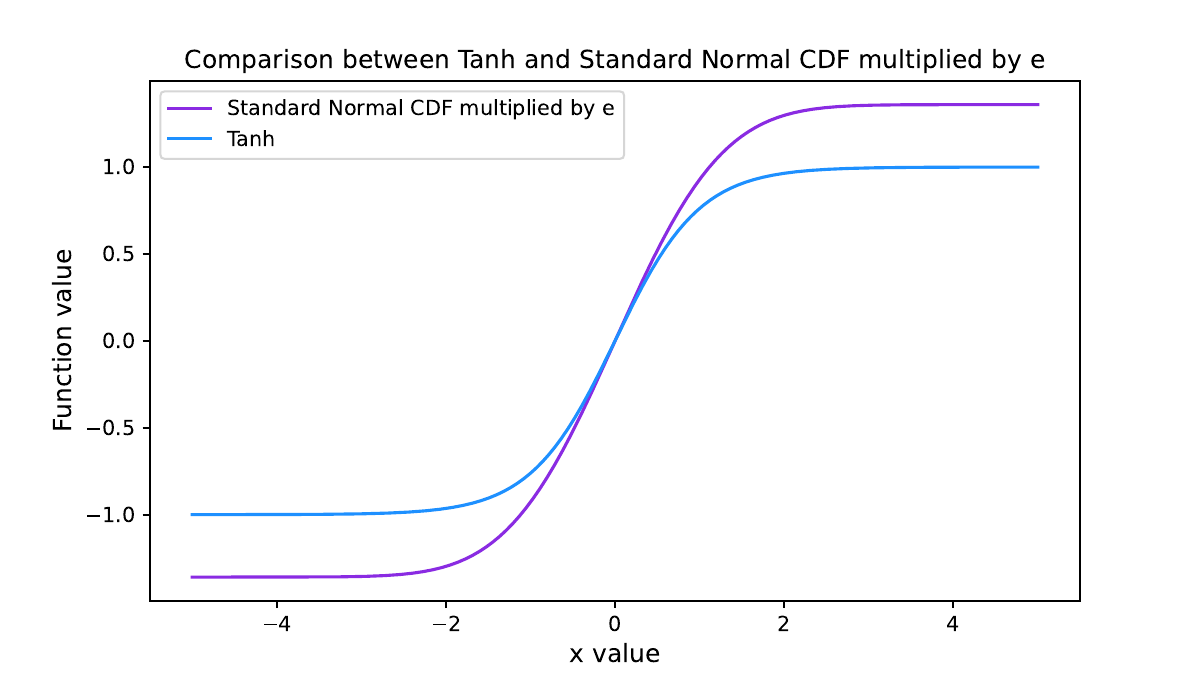}
    \vspace{-2em}
    \caption{Comparison between Tanh and Standard Normal CDF multiplied by $e$ (has been transformed to achieve symmetry about origin)}
    \label{Comparison between Tanh and Standard Normal CDF multiplied by e}
\end{figure}

Let's take several typical boundary conditions into consideration. Firstly, setting $f(x)$ approaches 1, when $x$ tends to positive infinity; and $f(x)$ approaches 0, when $x$ tends to negative
infinity. Therefore, the solution takes the form of cumulative distribution function (CDF), which can be expresses as:
\[
f(x)={\int_{-\infty}^x p(t)\text dt}
\]
Similarly, if fixing the difference between the upper and lower bounds of the activation function to be $e$, and making the activation function symmetric about the origin, the form can be written as:
\[
f(x)=e{\int_{0}^x p(t)\text dt}
\]
Furthermore, in the event that the input data distribution is assumed to be approximately uniformly distributed, the worst activation function can be approximated as a linear function. Were it to approximate the input data distribution as a normal distribution, then the form of the worst activation function would be closer to Sigmoid and Tanh. We show the comparison of function curves in Figure \ref{Comparison between Sigmoid and Standard Normal CDF} and Figure \ref{Comparison between Tanh and Standard Normal CDF multiplied by e}.

\section{Proof of Proposition \ref{proposition for CRReLU's inverse function}}
\label{proof of proposition for CRReLU's inverse function}
Before the proof of Proposition \ref{proposition for CRReLU's inverse function}, we would like to show four facts without proof.
\begin{fact}
\label{fact1}
    $f(x)=xe^{- \frac{x^2}{2}}$ is a bounded function, and range of the function is $[-e^{-\frac{1}{2}} , e^{-\frac{1}{2}}]$. 
\end{fact}
\begin{fact} 
\label{fact2}
    $f(x)=x^2e^{- x^2}$ is a bounded function, and range of the function is $[0 , e^{-1}]$. 
\end{fact} 
\begin{fact}
\label{fact3}
    $f(x)=x^3e^{-\frac{3}{2} x^2}$ is a bounded function, and range of the function is $[-e^{-\frac{3}{2}} , e^{-\frac{3}{2}}]$.
\end{fact}
\begin{fact}
\label{fact4}
    $\forall x \in \mathcal{R}$, $1-e^{-x}-x \leqslant  0$.
\end{fact}

We now commence the proof of Proposition \ref{proposition for CRReLU's inverse function}.
\begin{proof}
Substituting the analytic expression into the formula and performing algebraic simplifications, we can obtain:
\[
\begin{split}
    g\left(f(x)\right)=g\left(x+\epsilon xe^{-\frac{x^2}{2}}\right)&=x+\epsilon xe^{-\frac{x^2}{2}} - \epsilon \left(x+\epsilon xe^{-\frac{x^2}{2}}\right)e^{-\frac{1}{2}\left(x+\epsilon xe^{-\frac{x^2}{2}}\right)^2}\\
&=x + \epsilon x \left(e^{-\frac{x^2}{2}} - e^{-\frac{1}{2}\left(x+\epsilon xe^{-\frac{x^2}{2}}\right)^2}\right) - \epsilon ^2xe^{-\frac{x^2}{2}}e^{-\frac{1}{2}\left(x+\epsilon xe^{-\frac{x^2}{2}}\right)^2}  \\
&=x+\epsilon x e^{-\frac{x^2}{2}}\left[ 1-e^{-\frac{1}{2}\left(2\epsilon xe^{-\frac{x^2}{2}} + \epsilon^2x^2e^{-x^2}\right)}\right]-\epsilon^2xe^{-\frac{x^2}{2}}e^{-\frac{1}{2}\left(x+\epsilon xe^{-\frac{x^2}{2}}\right)^2} 
    \end{split}
\]
Thus, 
\[
\begin{split}
\left|g(f(x))-x\right|&=\left|\epsilon x e^{-\frac{x^2}{2}}\left[ 1-e^{-\frac{1}{2}\left(2\epsilon xe^{-\frac{x^2}{2}} + \epsilon^2x^2e^{-x^2}\right)}\right] - \epsilon^2xe^{-\frac{x^2}{2}}e^{-\frac{1}{2}\left(x+\epsilon xe^{-\frac{x^2}{2}}\right)^2} \right| \\
&\leqslant \left|\epsilon x e^{-\frac{x^2}{2}}\left[ 1-e^{-\frac{1}{2}\left(2\epsilon xe^{-\frac{x^2}{2}} + \epsilon^2x^2e^{-x^2}\right)}\right]\right|  \\
&\leqslant \left|\epsilon x e^{-\frac{x^2}{2}}\left[-\frac{2\epsilon xe^{-\frac{x^2}{2}} + \epsilon^2x^2e^{-x^2}}{2}\right]\right| \\
&= \left|\epsilon x e^{-\frac{x^2}{2}}\left(-\epsilon x e^{-\frac{x^2}{2}}-\frac{1}{2}\epsilon^2x^2e^{-x^2}\right)\right|=\left|\epsilon^2x^2e^{-x^2}+\frac{1}{2}\epsilon^3x^3e^{-\frac{3}{2}x^2}\right|\\
&\leqslant \left|e^{-1}\epsilon^2 + 0.5e^{-\frac{3}{2}}\epsilon^3\right|
\end{split}
\]
The first inequality is established owing to Fact \ref{fact1} and the fact that when $x$ is positive, the second term of absolute value must be positive. The second inequality is established owing to Fact \ref{fact4}. The third inequality is established owing to Fact \ref{fact2} and Fact \ref{fact3}. Hence, we can draw the conclusion that the absolute value of error between $g\left(f(x)\right)$ and $x$ is bounded with $\left|e^{-1}\epsilon^2 + 0.5e^{-\frac{3}{2}}\epsilon^3\right|$, which completes the proof.

\end{proof}

\section{Further details of CRReLU}
\label{Further details of CRReLU}
\subsection{Correction Regularized ReLU (CRReLU) Pseudocode}
\label{Correction Regularized ReLU (CRReLU) Pseudocode}

\begin{algorithm}[h]
\caption{Correction Regularized ReLU (CRReLU) Pseudocode}
\label{alg:pseudocode}
\lstset{
    basicstyle=\fontsize{9pt}{9pt}\ttfamily\bfseries,
    commentstyle=\fontsize{9pt}{9pt}\color{codeblue}
}
\begin{lstlisting}[language=Python]
import torch
import torch.nn as nn
import torch.nn.functional as F

class CRReLU(nn.Module):
    def __init__(self,lr=0.01):
        super(CRReLU,self).__init__()
        self.lr = nn.Parameter(torch.tensor(lr))
        
    def forward(self,x):
        return F.relu(x)+self.lr*x*torch.exp(-x**2/2)
\end{lstlisting}
\end{algorithm}

\subsection{Further Discussion on Properties of CRReLU}
\label{Further Discussion on Properties of CRReLU}
\begin{figure}[h]
    \includegraphics[width=1\linewidth]{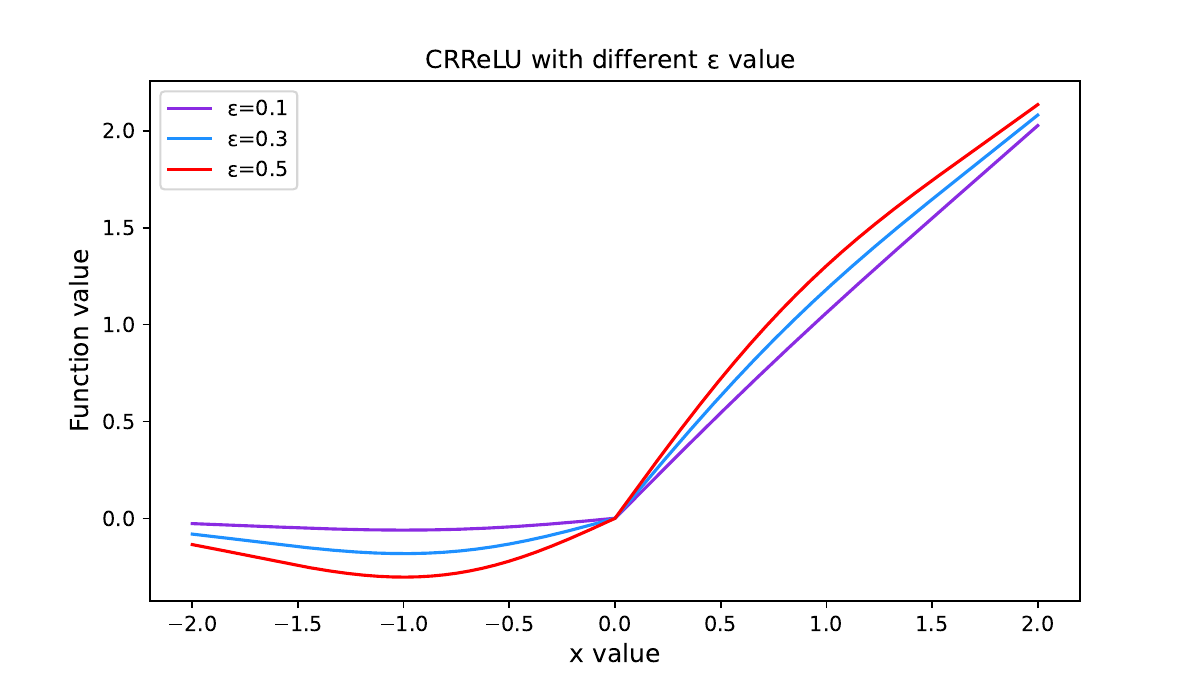}
    \caption{CRReLU with different $\epsilon$ value}
    \label{CRReLU with different ε value}
\end{figure}

We show the function curves with different $\epsilon$ values for CRReLU in Figure \ref{CRReLU with different ε value}. As depicted in the figure, existence of the correction term in CRReLU brings several good properties. It allows propagation of gradient when input is less than zero, serving to alleviate the dying ReLU phenomenon to a certain degree; simultaneously, as $x$ approaches negative infinity, CRReLU also converges to $0$, thereby guaranteeing sparsity of models in the negative part.

\section{Details of experimental settings}
\label{Details of experimental setting}
\subsection{Task of Image Classification}

\begin{table}[h]
\centering
\caption{Experimental settings of ViT, DeiT and TNT on CIFAR-10 and CIFAR-100 datasets}
\begin{tabular}{c|c}
\toprule
Image Size & 32 $\times$ 32 \\   \midrule
Patch Size & 4 \\       \midrule
Embedding Dim & 192 for ViT-Tiny and DeiT-Tiny ; 384 for TNT-small  \\       \midrule
Optimizer &AdamW with weight decay = 0.05  \\ \midrule
Learning Rate& \makecell{Cosine Annealing Learning Rate Scheduler \\
Initial lr = 2.5$\times 10^{-4}$ ; lr drop = -1 ; min lr = 1 $\times 10^{-5}$} \\ \midrule
Warm up& warmup epochs = 20 ; warmup learning rate = 1$\times10^{-6}$ \\  \midrule
Gradient Clipping  &  1.0 \\  \midrule
Training Epochs& 100 \\  \midrule
Batch Size&256\\  \midrule
Loss Function&CrossEntropy Loss\\  \midrule
Normalization & Layer Norm \\  \midrule
Data Augmentation& True (provided by timm)\\  \midrule
Drop Out and Drop Path & False\\
\bottomrule
\end{tabular}

\end{table}

\begin{table}[h]
\centering
\caption{Experimental settings of ViT and DeiT on ImageNet-1K dataset}
\begin{tabular}{c|c}
\toprule
Image Size & 224 $\times$ 224 \\   \midrule
Patch Size & 16 \\       \midrule
Embedding Dim & 192     \\       \midrule
Optimizer &AdamW with weight decay = 0.05  \\ \midrule
Learning Rate& \makecell{Cosine Annealing Learning Rate Scheduler \\
Initial lr = 2.5$\times 10^{-4}$ ; lr drop = -1 ; min lr = 1 $\times 10^{-5}$} \\ \midrule
Warm up& warmup epochs = 20 ; warmup learning rate = 1$\times10^{-6}$ \\  \midrule
Gradient Clipping  &  1.0 \\  \midrule
Training Epochs& 100 \\  \midrule
Batch Size&256\\  \midrule
Loss Function&CrossEntropy Loss\\  \midrule
Normalization & Layer Norm \\  \midrule
Data Augmentation& True (provided by timm)\\  \midrule
Drop Out and Drop Path & False\\
\bottomrule
\end{tabular}
\end{table}

\begin{table}[h]
\centering
\caption{We record changes in parameter number when employing various activation functions. GELU, ELU, CELU, SiLU (Swish), and Mish are considered activation functions without learnable parameter \textbf{(AFs without LP)}, while PReLU and CRReLU are considered activation functions with learnable parameter \textbf{(AFs with LP).} The results demonstrate that increase in parameter number introduced by the learnable parameter is negligible.}
\label{changes in parameter number}
\begin{tabular}{cc|ccc}
\toprule
\multicolumn{2}{c|}{Parameter Number}                            & CIFAR-10 & CIFAR-100 & ImageNet-1K \\ \midrule
\multicolumn{1}{c|}{\multirow{2}{*}{ViT-Tiny}}  & AFs without LP &  5399818  &  5417188  &   5754472  \\
\multicolumn{1}{c|}{}                           & AFs with LP    &   5399830 &   5417200  &  5754484   \\ \midrule
\multicolumn{1}{c|}{\multirow{2}{*}{DeiT-Tiny}} & AFs without LP &  5365076  &   5399816 &   5910800    \\
\multicolumn{1}{c|}{}                           & AFs with LP    &  5365088  &   5399828 &   5910812   \\ \midrule
\multicolumn{1}{c|}{\multirow{2}{*}{TNT-Small}} & AFs without LP &  21525298  &  21559948 &     /        \\
\multicolumn{1}{c|}{}                           & AFs with LP    &  21525322 &   21559972 &     /        \\ \midrule
\end{tabular}
\end{table}

\subsection{Task of Large Language Model (LLM) Fine-tuning}

\begin{table}[h]
\centering
\caption{Experimental settings of GPT2 fine-tuning task}
\begin{tabular}{c|c}
\toprule
Batch Size & 32     \\       \midrule
Optimizer& RMSprop (More Memory-Efficient)     \\ \midrule
Learning Rate& 5$ \times 10^{-7}$ with linear warmup steps of 150\\ \midrule
Trainer & FSDPTrainer (2 GPUs) \\ \midrule
Max Gradient Norm  &  10.0 \\  \midrule
Max Length for an Input (Prompt + Response)  &  512 \\  \midrule
Max Length for Prompt  &  256 \\  
\bottomrule
\end{tabular}
\end{table}

\end{document}